\pdfoutput=1

\documentclass[11pt]{article}

\usepackage[preprint]{acl}

\usepackage{times}
\usepackage{latexsym}
\usepackage{comment}

\usepackage[T1]{fontenc}

\usepackage[utf8]{inputenc}

\usepackage{microtype}

\usepackage{inconsolata}

\usepackage{graphicx}

\usepackage{amsmath}
\usepackage{comment}
\usepackage{amsmath}
\usepackage{pifont}
\usepackage{caption, booktabs}
\usepackage{algorithm}
\usepackage{algpseudocode}
\usepackage{amssymb}
\usepackage{setspace}
\usepackage{bbm}
\usepackage{arydshln}

\makeatletter
\newcommand\footnoteref[1]{\protected@xdef\@thefnmark{\ref{#1}}\@footnotemark}
\makeatother
\newcolumntype{P}[1]{>{\centering\arraybackslash}p{#1}}

\usepackage{xspace}
\newcommand{\ours}[0]{\textsc{MoI}\xspace}
\newcommand{\ourslong}[0]{Mixture-of-Intervention\xspace}

\usepackage{graphicx}
\usepackage{adjustbox}
\usepackage{subfig}
\makeatletter
\newcommand{\thickhline}{
    \noalign {\ifnum 0=`}\fi \hrule height 1pt
    \futurelet \reserved@a \@xhline
}
\newcolumntype{"}{@{\hskip\tabcolsep\vrule width 1pt\hskip\tabcolsep}}
\makeatother
\usepackage{mathabx}
\usepackage{amsfonts}

\makeatletter
\newcommand*{\blackleq}{
  \mathrel{
    \mathpalette\@blackleq{}
  }
}
\newcommand*{\@blackleq}[2]{
  \vcenter{
    \m@th
    \setbox0=\hbox{$#1\mkern3mu$}
    \setbox2=\hbox{$#1\vcenter{}$}
    \setbox4=\hbox{\raisebox{-\ht2}[.2pt][.2pt]{$#1-$}}
    \hbox{$#1\blacktriangleleft$}
    \nointerlineskip
    \kern\wd0 
    \copy4 
  }
}
\makeatother
\usepackage{dsfont}

\usepackage{multirow}
\usepackage{mathtools}

\usepackage{tablefootnote}

\usepackage{placeins} %

\usepackage{tikz}

\title{Inference Scaling for Bridging Retrieval and Augmented Generation}

\author{\textbf{
Youngwon Lee\textsuperscript{*}\quad
Seung-won Hwang\thanks{Work done while visiting Snowflake. Correspondence to: \href{mailto:seungwonh@snu.ac.kr}{\texttt{seungwonh@snu.ac.kr}}.}\quad
Daniel Campos}\\
\textbf{
Filip Graliński\quad
Zhewei Yao\quad
Yuxiong He
}\\
Snowflake AI Research\qquad Seoul National University\textsuperscript{\rm *}\\
}

\begin{document}
\maketitle

\begin{abstract}

Retrieval-augmented generation (RAG) has emerged as a popular approach to steering  the output of a large language model (LLM) by incorporating retrieved contexts as inputs.
However, existing work observed the generator bias,
such that
improving the retrieval results
may negatively affect the outcome.
In this work, we show
such bias can be mitigated,
from inference scaling, aggregating inference calls from the permuted order of retrieved contexts.
The proposed \ourslong (\ours) 
explicitly models the debiased utility of each passage with multiple forward passes to construct a new ranking.
We also show that
\ours can leverage the retriever's prior knowledge to reduce the computational cost by minimizing the number of permutations considered and lowering the cost per LLM call.
We showcase the effectiveness of \ours on diverse RAG tasks, improving ROUGE-L on MS MARCO and EM on HotpotQA benchmarks by $\sim 7$ points.

\end{abstract}

\section{Introduction}
\label{sec:intro}

\begin{figure}[t]
  \centering
  \includegraphics[width=.85\columnwidth]{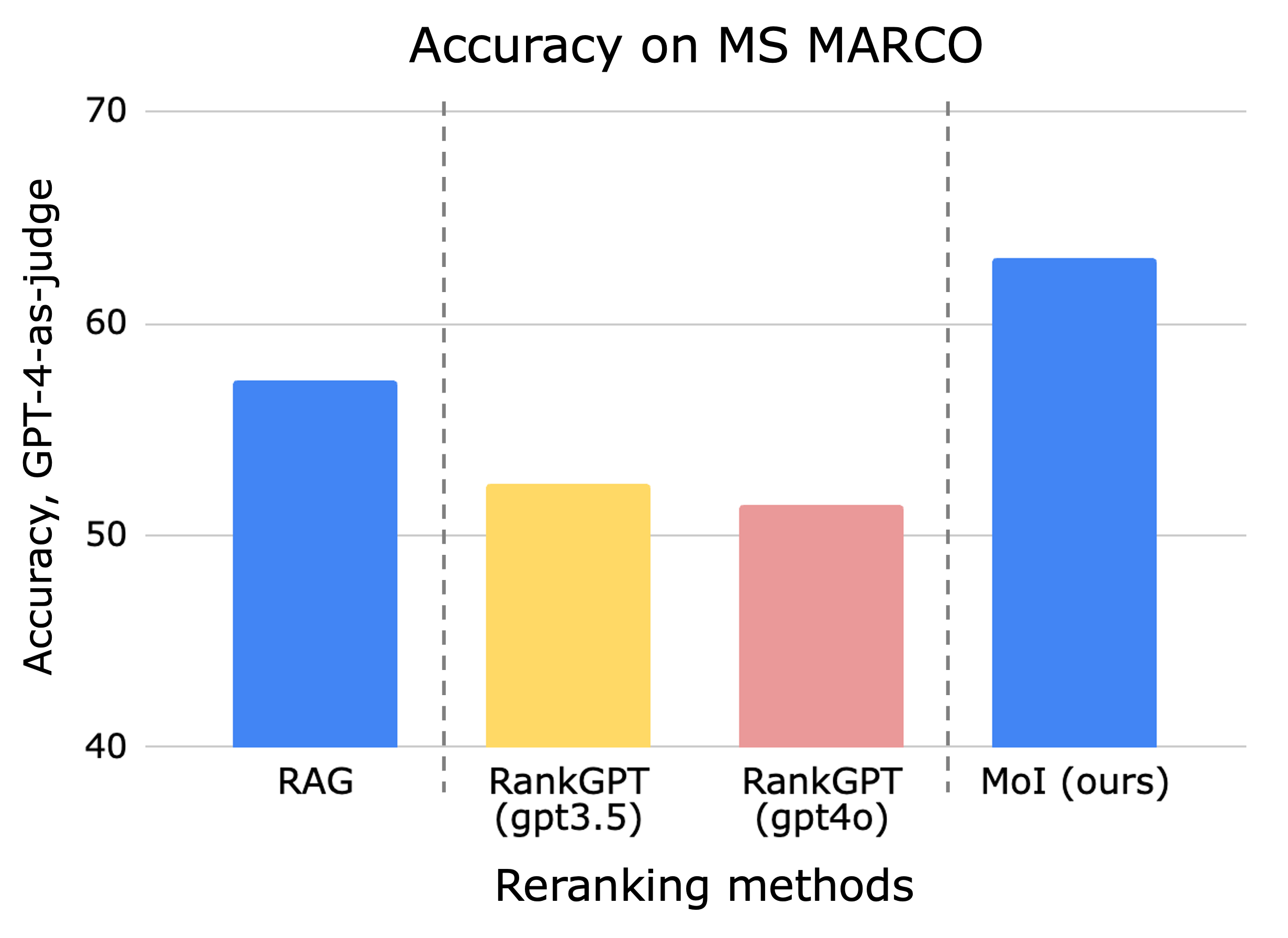}
  \caption{
  (Left, RAG) Top-10 passages retrieved by a complex retrieval system involving the Bing search engine are fed to the generator.
  (Center) RankGPT, a strong reranker based on LLM, hurts the performance, even more severely with stronger backbone.
  (Right) \ours improves the answer quality, outperforming RAG without reranking by a large margin of 6 points in accuracy.
  }
  \label{fig:rankgpt_fail}
\end{figure}

Retrieval-Augmented Generation (RAG) has become a widely adopted strategy to address core limitations of large language models (LLMs), such as hallucinations or restricted generalization to topics, concepts, or ideas that were not covered during training, by presenting relevant information to ground generation~\cite {gao2023retrievalaugmented}.

\begin{figure*}[t]
  \centering
  \includegraphics[width=\linewidth]{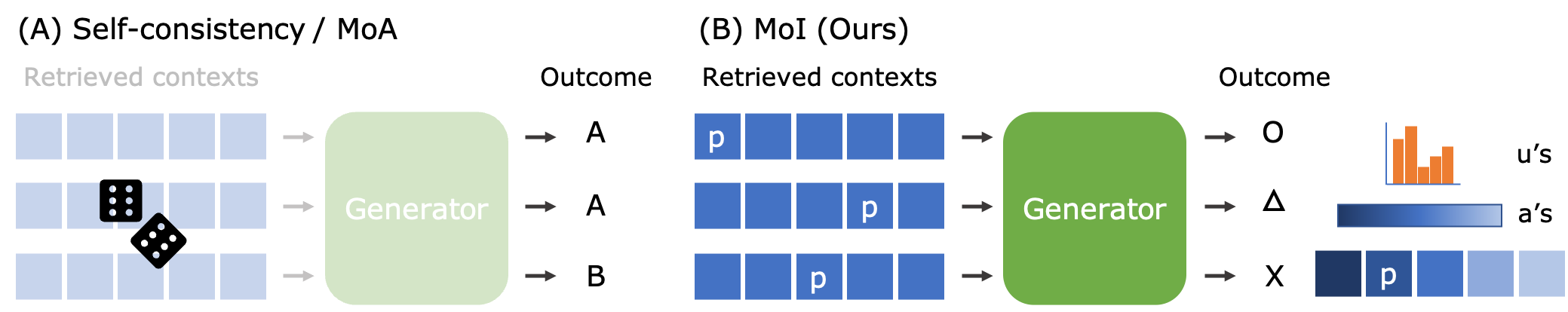}
  \caption{
  (A, baseline) Self-consistency~\citep{selfconsistency} and MoA~\citep{wang-etal-2024-mixture} treat random permutations of passages as black-box and count the consistency vote for outcomes.
  (B, proposed) In \ours, permutations are treated as white-box  intervention of one another, such that,
  from the obserevations of $p$ in varying positions, \ours estimates the effect of each passage on generation $u$ along the impact of position bias $a$.
  Finally, the ordering based on debiased utility $u$ is used for generation.
  }
  \label{fig:comparison}
\end{figure*}

However, existing work observed the generator bias,
such that
improving the retrieval results
may negatively affect the outcome.
As a bridge,
Figure~\ref{fig:rankgpt_fail}
demonstrates the use of reranker: However, RankGPT~\citep{sun-etal-2023-chatgpt}, a widely adopted reranker based on prompting LLMs, improves the retrieval quality but negatively impacts RAG performance on the MS MARCO benchmark~\citep{bajaj-etal-2018-ms}.
Even worse, employing a stronger backbone LLM for reranking worsens the quality further.
These unexpected results suggest that R's objective of maximizing relevance may not always
produce optimal outputs.
Meanwhile, training a dedicated bridge module for bridging such gap has been studied~\citep{ke-etal-2024-bridging}, which requires costly heuristic-based annotation to build the train set.

An alternative train-free approach is inference scaling, by aggregating generation from the \emph{permutations} of the retrieved results. This strategy, known as self-consistency~\citep{selfconsistency}, uses the number of permutations with consistent generation as a proxy for quality.
We call this baseline, a Mixture-of-Agents~(\citet{wang-etal-2024-mixture}; MoA) baseline:
Figure~\ref{fig:comparison}A depicts
MoA aggregates \textbf{blackbox} outputs from parallel, independent agent calls to AG, each fed with differently permuted retrieved results, to choose the output A, which is more consistently supported.

Unlike MoA, which uses multiple calls solely for consistency voting, we leverage these calls to observe the same passage in varying positions. This allows us to directly capture \textbf{position bias}—the LLM's disproportionate weighting of input contexts based on their relative position.

Specifically, \ours distinguishes two key factors: the true utility of each passage ($u$) and the effect of position bias ($a$), enabling debiased re-ranking of retrieved contexts.

For example, Figure~\ref{fig:comparison}B visualizes how we predicted bias: darker colors in $a$ represent stronger attention to passages in front positions, indicating $p$'s contribution to the outcome is overemphasized due to its position. This aligns with the ``lost-in-the-middle'' bias observed in prior work~\cite{liu-etal-2024-lost}. The final ranking, adjusted by debiased utility, moves 
$p$ to second position. Empirically, this reranking leads to the generator producing higher-quality answers.

Our contributions address the following research questions through the development of \ours:  
\begin{itemize}
    \item \textbf{(RQ1)} How can the debiased utility and ranking be determined from multiple inference calls?  
    \item \textbf{(RQ2)} Can \ours match the effectiveness of black-box MoA scaling, which requires inference calls for all permutations, while using fewer observations?  
    \item \textbf{(RQ3)} How can we reduce the inference cost per call, for example, by using a smaller model or input?  
\end{itemize}

In summary, our contribution is as follows:
\begin{itemize}

\item We demonstrate that enhancing the retriever or generator alone may not improve RAG, thereby highlighting the need for the bridge.
\item We propose a method to intervene in the ordering of retrieved contexts by explicitly modeling LLM position bias and aggregating diverse observations.
\item We show that the ranking determined by \ours improves downstream RAG task performance, leveraging retriever prior for efficient and effective intervention.
\end{itemize}

\section{Related Work}

This section overviews existing work on the bias mitigation in RAG.

\subsection{Mitigating Bias in RAG}

Our observation of RAG bias in Figure 1 was consistently made in ~\citep{izacard-2023-atlas,lin-2023-radit,izacard-grave-2021-leveraging}.
claiming the improved retrieval may not improve
RAG.
A widely adopted explanation is
position bias, also known as ``lost-in-the-middle''~\cite{liu-etal-2024-lost} problem, of the generator considering the passage in the middle less significantly.

\paragraph{Modifying the generator}
For dealing with such bias, a common approach has been updating the generator LLM, often jointly trained with the retriever as well~\citep{izacard-2023-atlas}.
Alternatively,
positional embeddings and attention matrices have been manipulated to debias~\citep{wang-etal-2024-eliminating,ratner-etal-2023-parallel}, often aiming for complete order invariance.
However, as
LLMs were never exposed to such manipulated embeddings or attention weights/masks during training, 
they may suffer from unexpected degradation in performance, such as multi-hop reasoning capabilities~\citep{yang-2023-revisiting}.
Recently, \citet{hsieh-etal-2024-found} also studied modifying the generator side, using the average attention weights assigned to passages to detect and account for bias.

\paragraph{Training bridge}
Among  solutions, our work is most closely related to \citet{ke-etal-2024-bridging},  training a `bridge' model between the retriever and generator,
 by selecting an ordered subset of retrieved passages.

\paragraph{Blackbox inference scaling}
When retraining retriever or generator, or jointly both is not feasible,
a widespread approach is to rely on inference-time scaling, such as
self-consistency~\citep{selfconsistency} or Minimum Bayes-Risk decoding~\citep{kumar-byrne-2004-minimum} mechanism.
For example, \citet{tang-etal-2024-found} have generated several hypothesis rankings from different permutations of passages as inputs and then selected the one closest to other rankings in IR reranking task.

\paragraph{Our distinction} 

Our method is whitebox inference scaling that can be interpreted as implementing  bridge mechanism without training a separate bridge module, while leaving the retriever and generator intact.
As such, our work is orthogonal to improving the retriever or generator,
which can be combined with those approaches.

\subsection{Mitigation by Mixture of Agents}

As Figure 2A illustrates,
self-consistency~\citep{selfconsistency}
over permuted orders can
mitigate bias by marginalizing the latent variables.
Under a similar setting to ours, \citet{tang-etal-2024-found} used self-consistency mechanism to account for the position bias for IR reranking task.

We build a MoA baseline~\citep{wang-etal-2024-mixture}.
where several LLM agents are called in parallel to independently generate an output given the same input, 
to hide inference latency of multiple calls.
This corresponds to two phases:
first \emph{propose} phase generating output from permuted orders, and then \emph{aggregate} to produce the final single reranked sequence of contexts.

\paragraph{Our distinction}
We view permutations as the intervention 
of one another, allowing \textbf{strategized proposal} phase,
followed by \textbf{efficient aggregation}, where the cost of inference call is further reduced.

\section{Method}
\label{sec:method}

\subsection{Overview}
Our proposed method, dubbed \ourslong (\ours), disentagles the \emph{utility} $u$ of each retrieved context, from the effect of \emph{position bias} $a$ to the given
generator,
shown by color gradations in Figure~\ref{fig:comparison}B.
To better explain how \ours simultaneously computes both and why this is crucial, we first review how previous works obtain utility alone.

For instance, the Bayesian saliency score~\citep{merth2024superposition,muennighoff-2022-sgpt} defines the following pointwise score: 
\begin{equation} \label{eq:bayes_score}
u_p := P(p \,|\, q)\ \propto\ P(q \,|\, p) P(p)
\end{equation}
derived from probabilities given by the generator LLM. This score measures the saliency of passage
 passage \( p \) relative to query $q$.
Note that dropping the second term $P(p)$ 
results in a variant used in question generation (QG)~\citep{sachan-etal-2022-improving}, which estimates how likely
 $q$ would be answered by $p$.

However, this approach fails to account for how multiple passages collaborate in answer generation, for which,
Eq.~\ref{eq:bayes_score} can be generalized to a listwise score
\begin{multline} \label{eq:bayes_score_iterative}
u_p = P(p \,|\, q, p_1, \cdots, p_k)\ \propto \\
\hfill P(q \,|\, p_1; \cdots; p_k; p) P(p \,|\, p_1; \cdots; p_N).
\end{multline}
where $p_1$ through $p_k$ denote the $k$ passages 
that have been \emph{sequentially} selected with the passage with the highest listwise $u_p$ score.

While this approach enables to
model collaborative utility of $p$
to other passages in the list,
it has two shortcomings.
The sequential nature of modeling listwise effect, requires $\mathcal{O}(N^2)$ number of evaluations of $u_p$ which incurs $\mathcal{O}(N)$ latency even when provided with enough compute to parallelize.
Another shortcoming is that it cannot
 observe how $u_p$ changes when different passages were selected before, also due to the sequential dependencies.

\ours breaks dependency by
observing $p$ from diverse context,
applying interventions of orders independently in parallel.
These parallel observations enable
to disentangle utility from positional bias, by aggregating the outcomes from different permutations of the passages, thus allowing the model to observe how varying the order of the passages influences the generation. 

Formally, given a set of $N$ retrieved passages $\{p_1, \cdots, p_N\}$ deemed relevant to a query $q$ and $M$ permutations $\pi_1, \cdots, \pi_M$ over $1, 2, \cdots, N$, we define and observe the outcome of a permutation $\pi_i$,
\begin{multline} \label{eq:ours}
s_i = P\left( q \,|\, p_{\pi_i[1]}; p_{\pi_i[2]}; \cdots ; p_{\pi_i[N]} \right)\\
\hfill \times P\left( p_{\pi_i[1]}; p_{\pi_i[2]}; \cdots ; p_{\pi_i[N]} \right),
\end{multline}
where $\pi_i[j]$ denotes the index of the passage placed at $j$-th position according to the $i$-th permutation $\pi_i$.
We can see from definition that $s_i$, depends on (1) what other passages are in the prompt, and (2) how they are ordered in $\pi_i$.

\begin{figure}[t]
  \centering
  \includegraphics[width=\linewidth]{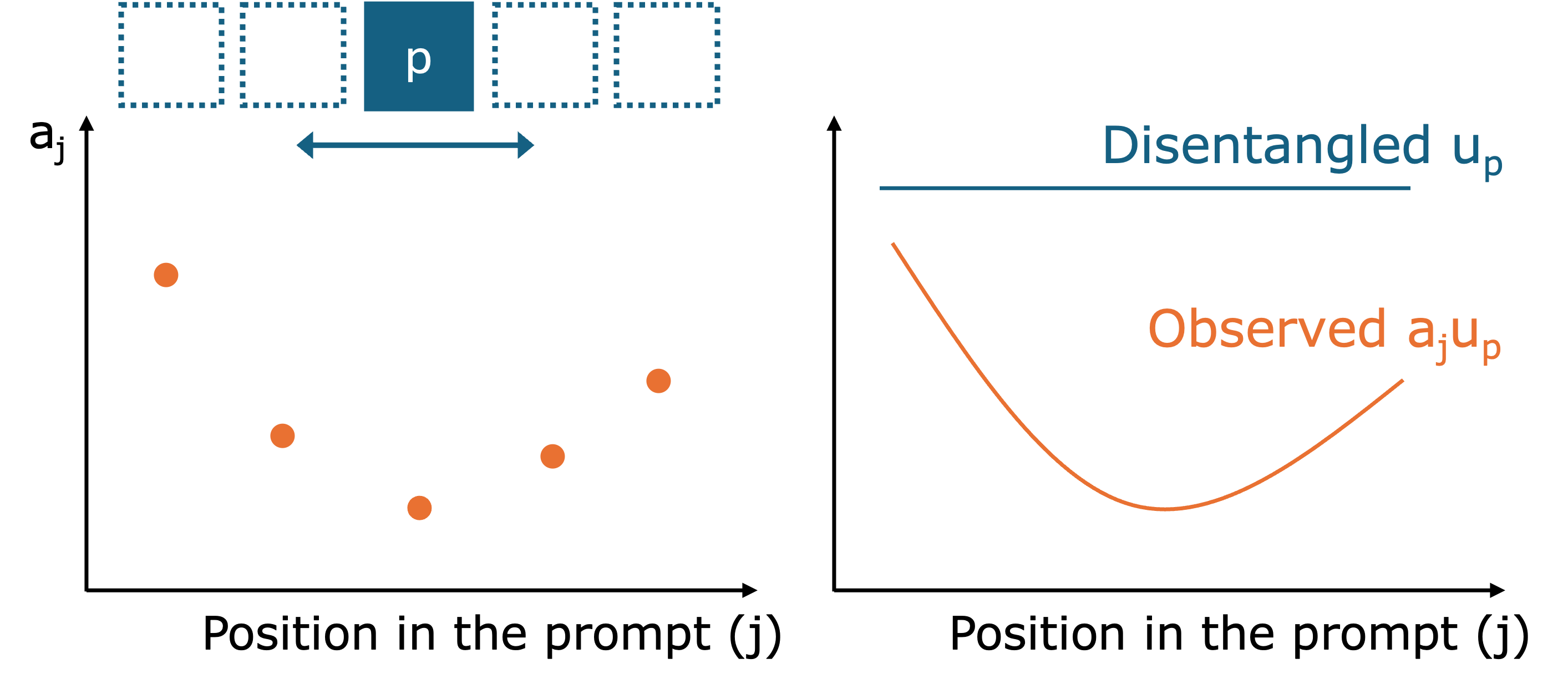}
  \caption{
  Ideally, wherever a passage $p$ is placed, its contribution to generation, or utility, should be constant (blue line).
  However, due to position bias of LLMs, the observed orange curve varies by the position and surrounding context.
  \ours disentangles the effect of position bias (left figure) from observation, to determine the debiased utility $u_p$ through multiple parallel interventions.
  }
  \label{fig:u_and_p}
\end{figure}

We aim to disentangle listwise scores $s_i$ into 
into two components: utility and position bias. To model this, we introduce positional 
bias $a_j$'s 
to well predict $s_i$ as a weighted sum for each permutation $\pi_i$: 
\begin{equation} \label{eq:u_and_p_sum}
\sum_{1\leq j\leq N} a_j \cdot u_{\pi_i[j]}
\end{equation}

Figure~\ref{fig:u_and_p} illustrates this idea, where the position bias of the LLM makes the contribution of a passage $p$ in Eq~\ref{eq:u_and_p_sum}  vary by its relative position in the prompt.
In previous works such as \citet{liu-etal-2024-lost}, 
this effect was measured by moving a single gold passage to observe the outcome at different positions, while ignoring the order of other passages. 
\ours generalizes this idea by 
simultaneously determining the effects of position bias and the debiased utility
 $u_p$ of each passage $p$, based on 
 parallel observations from multiple passage permutations.
Rather than observing $a_ju_p$ for each $j$ and $p$, by moving $p$'s relative position in the prompt, \ours aggregates the outcomes to estimate $a_j$'s and $u_p$'s for all $j$ and $p$  in Eq~\ref{eq:u_and_p_sum}.

In practice, we solve for $u_p$
by minimizing the L2 loss between the predicted and observed outcomes, subject to the constraint that positional coefficients sum to 1, ensuring a valid bias distribution. Nonlinear programming solvers are used to efficiently find the optimal values for $a_j$ and $u_p$:
\begin{align*}
\textrm{minimize} & \sum_{1\leq i\leq M} \left( \textstyle\sum_{1\leq j\leq N} a_j \cdot u_{\pi_i[j]} - s_i \right)^2 \\
\textrm{subject to} & \textstyle\sum_j a_j = 1,\ 0 \leq a_j \leq 1.
\end{align*}

After obtaining the scores,
we reorder 
based on descending true utility $u_j$  and feed this sequence back to the generator LLM, completing the \ours pipeline.\footnote{Empirical overhead of calling solvers was roughly 3\% of the cost of a single forward pass on GPU, in terms of wall-clock time.}

\subsection{Strategized propose phase}
\label{sec:method_elaborate}

As contrasted in Section 2.2, we improve MoA in two phases: 
\textbf{proposing} permutations and  \textbf{aggregating} as black-box by consistency, into
 \textbf{strategized propose} phase, of selecting informative orderings, and \textbf{efficient aggregate} phase, which disentangles utility and bias from the outcomes. Below, we elaborate on the implementation and potential optimizations for each phase.

\subsubsection{Random samples}

One extreme approach is to aggregate the entire ``universe'' set $U$ of all $N!$ possible permutations. Instead, we propose randomly sampling a subset $S \subset U$, with $|S|=3N$.
This ensures that we have enough equations to solve for $2N$ variables (i.e., $N$ for the $u$'s and another $N$ for the $a$'s). Importantly, these calls can be executed in parallel, leading to an overall latency equivalent to a single call.

\subsubsection{Comprehensiveness in  sampling}

\label{subsubsec:reduce_number}

We aim to strategize sampling by selecting a smaller but more ``comprehensive''  $S$.

\begin{figure}[t]
  \centering
  \includegraphics[width=.95\linewidth]{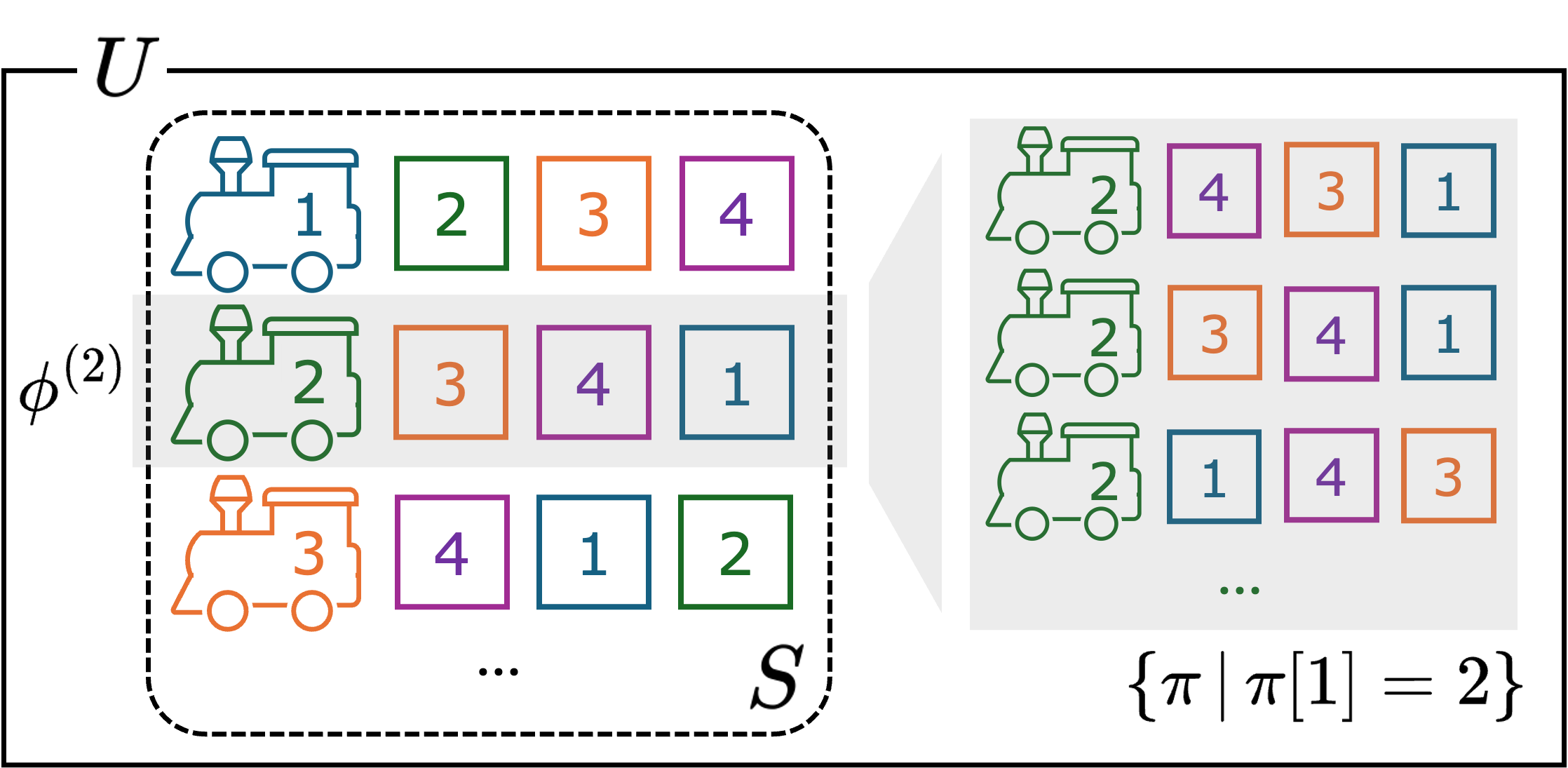}
  \caption{
  To approximate a comprehensive subset, we consider the set of cyclic permutations as $S$,
  encompassing diverse yet representative permutations to allow desirable ones to be surfaced.
  }
  \label{fig:comprehensive_subset}
\end{figure}

Ideally, if we could map any ordering outside $S$ to its ``counterpart'' in $S$—which better suits the generator’s preferences—then considering only $S$ would be comprehensive~\citep{hwang-chang-2007-optimizing}, or equally effective to consider the entire universe set $U$. We approximate this notion by ensuring $S$ to represent the broader landscape of $U$, as illustrated in Figure~\ref{fig:comprehensive_subset}.
Specifically, the shaded area indicates that permutations starting with passage 2 can be mapped to a representative permutation, $\phi^{(2)}$. 
We leave a formal definition of $\phi$
and an explanation on why $\phi^{(2)}$
can represent experiments
on shaded permutations starting with 2,
but the high-level intuition builds on a prior finding that the first element has the greatest influence on generation~\citep{hsieh-etal-2024-found,liu-etal-2024-lost}.

Formally, we propose to pick $S$ as the set of cyclic permutations where $|S|=N$.
Desirably, (1) each passage should have equal chance of being placed at each position in $S$, and
(2) each permutation in $S$ should represent distinct set of permutations in $U$, which would map to itself.
Figure~\ref{fig:comprehensive_subset} illustrates that our choice for $S$ achieves both criteria: (1) it chains passages in a round-robin fashion, and (2) it ensures even coverage of all the permutations in $U$.

\subsection{Efficient aggregate phase}
 
Next, we explore ways to reduce the cost of each call during the aggregate phase by (a) pruning the input contexts, and (b) utilizing a smaller distilled model, or SLM, instead of an LLM,   addressing our second research question.

\subsubsection{Smaller input to agent}

To cut down the cost of each call, we prune the input contexts by using the retriever's ranking as the reference ordering, $\phi^{(1)}$. This idea follows from \citet{reddy-etal-2024-first}, which demonstrated that the probability distribution over the first token accurately reflects the order intended by a reranker trained to generate passage ID sequences.

Rather than decoding the entire sequence, examining the model's prediction for the first passage ID significantly reduces costs. Similarly, we use the prefix containing the first $L<N$ passages of each permutation in $S$ to approximate the full $N$.
We denote the pruned permutation shifted by $k-1$ positions, in which $p_k$ is placed at first, as:
\begin{equation}
\phi_p^{(k)} = [p_{k}, p_{k+1}, \cdots, p_{k+L-1}],
\end{equation}
for $k\leq N-L$, or otherwise as
\begin{equation}
\phi_p^{(k)} = [p_k, \cdots. p_{N}, p_1, \cdots, p_{k+L-N-1}].
\end{equation}
This pruning strategy replaces the full permutation $\phi^{(k)}$ while still preserving the essential information for generation.

\begin{table*}[t]
    \centering
    \fontsize{10.5pt}{12pt}\selectfont
    \begin{tabular}{lrrrrrr}
        \thickhline
         & \multicolumn{2}{c}{\multirow{1.25}{*}{MS MARCO}}
         & \multicolumn{2}{c}{\multirow{1.25}{*}{HotpotQA}}
         & \multicolumn{2}{c}{\multirow{1.25}{*}{CRAG}}
         \\
        \cmidrule(r){2-3}
        \cmidrule(r){4-5}
        \cmidrule(r){6-7}
        \multirow{-1.25}{*}{Ranking} 
         & \multicolumn{1}{c}{\multirow{-1.25}{*}{R-L}}
         & \multicolumn{1}{c}{\multirow{-1.25}{*}{GPT-4}}
         & \multicolumn{1}{c}{\multirow{-1.25}{*}{EM}}
         & \multicolumn{1}{c}{\multirow{-1.25}{*}{GPT-4}}
         & \multicolumn{1}{c}{\multirow{-1.25}{*}{}}
         & \multicolumn{1}{c}{\multirow{-1.25}{*}{GPT-4}}
        \\\hline
        Retriever & 
            {37.75} & {57.28}
            & {$-$} & {$-$} 
            & {} & {52.43} 
            \\
        Random & 
            {35.92} & {51.56}
            & {48.54} & {73.88} 
            & {} & {51.94} 
            \\
        RankGPT~\citep{sun-etal-2023-chatgpt} & 
            {37.53} & {51.45}
            & {52.22} & {74.88} 
            & {} & {50.49} 
            \\
        Bayes saliency~\citep{merth2024superposition} & 
            {37.64} & {54.37}
            & {52.22} & {77.84} 
            & {} & {48.06}
            \\
        Bayes saliency +~\citep{merth2024superposition} & 
            {34.47} & {52.43}
            & {50.25} & {75.37} 
            & {} & {47.58}
            \\
        QG~\citep{sachan-etal-2023-questions} & 
            {36.78} & {55.34}
            & {48.28} & {73.89} 
            & {} & {53.40}
            \\
        LongLLMLingua~\citep{jiang-2023-longllmlingua} & 
            {33.21} & {50.49}
            & {49.75} & {74.39} 
            & {} & {50.49} 
            \\
        Self-consistency & 
            {38.45} & {58.26}
            & {51.72} & {76.85} 
            & {} & {54.37} 
            \\
        \rowcolor{gray!10} \ours (Ours) & 
            {\textbf{44.30}} & {\textbf{63.11}}
            & \textbf{55.67} & \textbf{79.81} 
            & {} & {\textbf{59.23}} 
            \\
        \thickhline
    \end{tabular}
    \caption{
    Results on different question answering benchmarks with LLaMa 3 8B as the generator and various reranking methods
    applied. For all metrics considered, higher the better.
    }
    \label{tab_main}
\end{table*}

\subsubsection{Smaller agent}
\label{subsubsec:small_model}

Another way to reduce the cost of each call is to delegate calls to a smaller model than the generator LLM.
For this purpose, we propose preference distillation, of turning a smaller agent to align 
and replacing LLM, thereby
featuring smaller memory and compute footprint.

\begin{figure}[t]
  \centering
  \includegraphics[width=\linewidth]{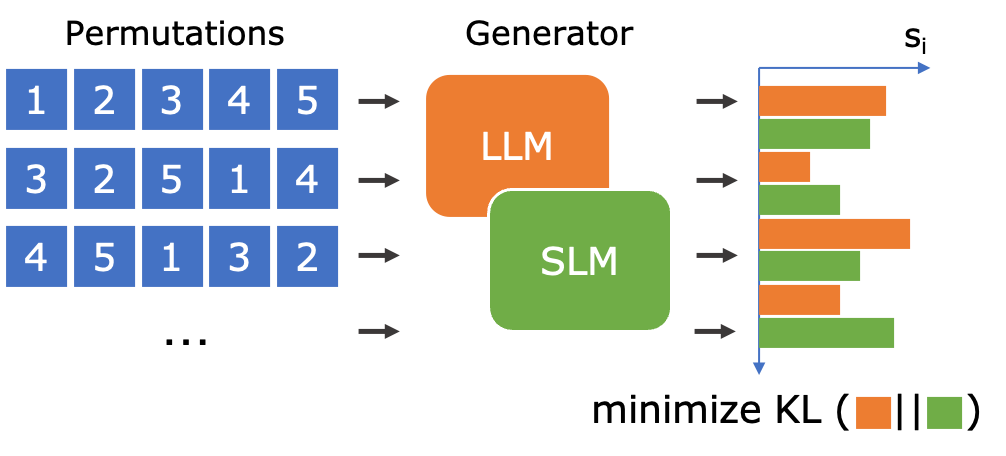}
  \caption{
  The distribution of $s_i$ from an LLM is distilled to a smaller model by minimizing KL between the normalized probability distributions after softmax.
  Values colored orange can be pre-computed.
  }
  \label{fig:distill}
\end{figure}

First, we compute permutation-wise saliency score defined in Eq~\ref{eq:ours} for $K$ random permutations of passages for each query using the LLM,
to construct an offline dataset.
During training, we randomly select $K^\prime$ permutations for each query and compute $\tilde{s_i}$'s using the small model.
For those $K^\prime$ permutations, softmax operation is applied to $s_i$'s and $\tilde{s_i}$'s to obtain probability distributions, and then the KL divergence between the two is minimized, as described in Figure~\ref{fig:distill}.

Our preference distillation enjoys the following advantages over training a bridge network
\citet{ke-etal-2024-bridging}, that learns to directly output the reranked sequence of subset of passages:
First, the train data preparation is much cheaper and easily parallelizable, than to repeat generating and evaluating the answer to iteratively build a pseudo-reference sequence.
Second, distillation exposes the model to dense supervisory signals, as opposed to presenting a single sequence per query as a positive demonstration, or, sparse supervision.
Our goal of distilling preference is more feasible than training a small model to directly output the desirable ranking, which eliminates additional round of reinforcement learning training as in \citet{ke-etal-2024-bridging}.

\section{Experimental Results}

\subsection{Experimental settings}
\label{subsec:exp_setting}

While we have mainly focused on question answering (QA) task, we also report results on other tasks, namely citation generation and fact verification.
For QA benchmarks, we employed the widely used MS MARCO dataset for single-hop reasoning scenarios, HotpotQA~\citep{yang-etal-2018-hotpotqa} for 2-hop reasoning, and CRAG~\citep{yang2024crag} for challenging multi-hop reasoning.
For citation generation and fact verification task, we used TREC-RAGgy and FEVER benchmarks, respectively.

For the backbone generator LLM, we used the publicly available LLaMA-3 and Phi-3 model families. Additionally, following prior work, we used greedy decoding to generate answers to ensure both efficiency and deterministic outputs.

For automatic evaluation of the generated answers, we adhered to the established evaluation protocols widely adopted for each benchmark. ROUGE-L~\citep{lin-och-2004-automatic} was used for MS MARCO, and exact match (EM) for HotpotQA, both of which are reference-based metrics that compare the predicted answers to ground-truth answers based on lexical overlap. We also employed GPT-4 for automatic evaluation, following \citet{yang2024crag}, which allowed us to assess answer quality more flexibly by accommodating responses with minor lexical variation while maintaining the core correctness of the answer.\footnote{While the original scoring from \citet{yang2024crag} outputs scores in the range of $[-100, 100]$, we rescale the score and report values in $[0, 100]$.}
To support this decision, in Appendix~\ref{app:humaneval}, we provide results from our user study, consistent with prior literature on LLM-as-a-judge, showing higher correlation with human judgment than traditional metrics.

\subsection{Effectiveness of \ours}

Table~\ref{tab_main} presents downstream performance of several reranking strategies on the question-answering task, highlighting the superior performance achieved by \ours. Rerankers generally exhibit poor performance, regardless of whether they model absolute relevance (e.g., RankGPT) or use signals from the generator. In contrast, self-consistency provides consistent performance improvements across benchmarks, though the gains are smaller compared to those from \ours.
We provide further qualitative analyses of the rankings determined by \ours and baselines in Appendix~\ref{app:qualitative}.

\begin{table}[t]
    \centering
    \begin{tabular}{lrr}
        \thickhline
         & \multicolumn{2}{c}{\multirow{1.25}{*}{TREC-RAGgy}}
         \\
        \cmidrule(r){2-3}
         \multirow{-1.25}{*}{Ranking} 
         & \multicolumn{1}{c}{\multirow{-1.25}{*}{FP}} & \multicolumn{1}{c}{\multirow{-1.25}{*}{FN}}
           \\\hline
        Retriever (BM25) & 
            {12.05} & {28.34}
            \\
        \rowcolor{gray!10} \ours (ours) & 
            {\textbf{11.40}} & {\textbf{24.76}}
            \\
        \thickhline
    \end{tabular}
    \caption{
    Percentage ratio of sentences with false positive (FP) and false negative (FN) citation errors on TREC-RAGgy dev set. Metrics are lower the better.
    }
    \label{tab_trec}
\end{table}

Additionally, we demonstrate that \ours can be applied beyond its role in question-answering systems to any RAG task.
To this end, we used LLaMA 3 8B as a citation generator to identify the passages supporting each sentence in a long-form response to a query on TREC RAGgy development set. 
Table~\ref{tab_trec} shows that the ordering of retrieved contexts (and how they are numbered for identification) also affects the output in this scenario, while \ours effectively reduces both types of errors.

We also observed consistent results on another knowledge-intensive task, fact verification, using the FEVER benchmark. Given the top-5 passages retrieved using DPR~\citep{karpukhin-etal-2020-dense} from Wikipedia, the generator was asked to classify the given statement as either true or false.
The accuracy reported in Table~\ref{tab:fever} again validates the effectiveness of our method across various tasks, outperforming baselines.

\begin{table}[t]
    \centering
    \begin{tabular}{lr}
        \thickhline
         Ranking
         & \multicolumn{1}{c}{Acc}
           \\\hline
        Retriever (DPR) & 
            {83.11}
            \\
        Random & 
            {83.42}
            \\
        RankGPT & 
            {83.88}
            \\
        Self-consistency & 84.07 \\
        \rowcolor{gray!10} \ours & \textbf{85.03} \\
        \thickhline
    \end{tabular}
    \caption{
    Fact verification performance on FEVER benchmark. We used the top-5 retrieved passages in \citet{wang-etal-2023-learning}.
    }
    \label{tab:fever}
\end{table}

\subsection{Cost-effective proposal and aggregation}
\label{subsec:cut_cost}

\paragraph{Model substitution}
To optimize the cost associated with intervention in \ours, we presented several designs in Section~\ref{sec:method_elaborate}.
We start by finding a balance between cost and performance through the use of a smaller substitute model as the agent.
Table~\ref{tab:off_the_shelf} demonstrates that replacing Phi-3 7B with an off-the-shelf Phi-3 3B retains 80\% of the performance gains over the random baseline, at approximately half the cost.
This can be attributed to that
models from the same family generally being pre-aligned and sharing similar preferences for passage permutations, as they are often trained on the same or very similar set of preference data.

\begin{table}[t]
    \centering
    \begin{tabular}{lrr}
        \thickhline
         & \multicolumn{2}{c}{\multirow{1.25}{*}{HotpotQA}}
         \\
        \cmidrule(r){2-3}
         \multirow{-1.25}{*}{Ranking} 
         & \multicolumn{1}{c}{\multirow{-1.25}{*}{EM}}
         & \multicolumn{1}{c}{\multirow{-1.25}{*}{GPT-4}}
           \\\hline
        \ours & 
            {55.67} & 79.81 %
            \\
        \phantom{00}+ replace w/ Phi-3 3B & 
            {54.18} & 78.82 %
            \\
        Random & 
            {48.36} & 71.64 %
            \\
        \thickhline
    \end{tabular}
    \caption{
    Replacing Phi-3 7B with Phi-3 3B
    cuts the cost nearly 50\% while 80\% of the performance improvement over the random baseline is maintained.
    }
    \label{tab:off_the_shelf}
\end{table}

\begin{table}[t]
    \centering
    \begin{tabular}{lrr}
        \thickhline
         & \multicolumn{2}{c}{\multirow{1.25}{*}{HotpotQA}}
         \\
        \cmidrule(r){2-3}
         \multirow{-1.25}{*}{Ranking} 
         & \multirow{-1.25}{*}{EM} & \multirow{-1.25}{*}{GPT-4}
           \\\hline
        \ours & 
            {55.67} & 79.81 %
            \\
        \phantom{0}+ replace w/ Phi-3 3B & 
            {49.26} & 74.39 %
            \\
        \phantom{00}+ distillation & 
            {53.69} & 79.81 %
            \\
        Random & 
            {48.54} & 73.88 %
            \\
        \thickhline
    \end{tabular}
    \caption{
    Results on HotpotQA with LLaMA3 8B model as generator. While replacing it with Phi-3 3B is not effective, after preference distillation 70/100\% of the gain in terms of the two metrics over the random baseline can be retained at $\sim$40\% inference cost.
    }
    \label{tab:pref_distill}
\end{table}

\paragraph{Preference distillation}
If a smaller model is not readily available for a given LLM, a suitable one can be created through preference distillation.
Table~\ref{tab:pref_distill} shows that the Phi-3 3B model is not effective as a direct substitute for the LLaMA 3 8B generator.
However, after performing preference distillation, the Phi-3 3B model can achieve the same performance score in GPT-4 evaluations at around 40\% of the inference cost.
The training details are provided in Appendix~\ref{app:implementation}.

\begin{table}[t]
    \centering
    \begin{tabular}{lrr}
        \thickhline
         & \multicolumn{2}{c}{\multirow{1.25}{*}{CRAG}}
         \\
        \cmidrule(r){2-3}
         \multirow{-1.25}{*}{Ranking} 
         & \multirow{-1.25}{*}{GPT-4}
           \\\hline
        \ours & 
            59.23
            \\
        \phantom{0} + Propose from cyclic & 
            53.40
            \\
        \phantom{00} + Pruning & 
            54.37
            \\
        \phantom{00} + Variable Pruning &
        54.86
        \\ 
        Random &
        49.26
        \\
        \thickhline
    \end{tabular}
    \caption{
    Leveraging retriever prior in both reducing the number of calls and the cost of each call on CRAG with Phi-3 7B as generator.
    }
    \label{tab:prior_helps}
\end{table}

\paragraph{Retriever prior}
As discussed earlier, we leverage prior knowledge from the retriever for efficiency in two ways.
First, we consider cyclic permutations based on the retriever's ranking to reduce the number of calls.
Next, sequences are pruned to a shorter length, which reduces the cost of each call. In this process, if the scores from the retriever are also available, they can further enhance the outcome, as shown in Table~\ref{tab:prior_helps}.
By adopting cyclic permutations and fixed-length pruning, we achieved 90+\% cost savings while maintaining 50\% of the relative performance gains compared to the random baseline, while variable pruning with retriever scores provided additional improvements.

\section{Analysis}

\label{sec:analysis}

\begin{table}[t]
    \centering
    \begin{tabular}{lrr}
        \thickhline
         & \multicolumn{2}{c}{\multirow{1.25}{*}{MS MARCO}}
         \\
        \cmidrule(r){2-3}
         \multirow{-1.25}{*}{Ranking} 
         & \multirow{-1.25}{*}{MRR} & \multirow{-1.25}{*}{ROUGE-L}
           \\\hline
        Retriever (Bing) & 
            {.338} & {37.75}
            \\
        Bayes Saliency & .353 & 37.64 \\
        Question Generation & .435 & 36.78 \\
        RankGPT & \textbf{.634} & 37.53 \\
        \rowcolor{gray!10} \ours (ours) & 
            {.464} & {\textbf{44.30}}
            \\
        \hline
        Gold at 2nd & .500 & 40.27 \\
        Gold at 3rd & .333 & 39.69 \\
        Gold at 4th & .250 & 36.05 \\
        \thickhline
    \end{tabular}
    \caption{Retrieval performance measured by MRR and downstream RAG performance measured by ROUGE-L.
    \ours 
    outperforms others with similar or higher mean reciprocal rank, by strategically ranking the gold lower.
    }
    \label{tab:goldrank}
\end{table}

\begin{table}[t]
    \centering
    \begin{tabular}{lrr}
        \thickhline
         & \multicolumn{2}{c}{\multirow{1.25}{*}{MS MARCO}}
         \\
        \cmidrule(r){2-3}
         \multirow{-1.25}{*}{Ranking} 
         & \multicolumn{1}{c}{\multirow{-1.25}{*}{EM}} & \multicolumn{1}{c}{\multirow{-1.25}{*}{GPT-4}}
           \\\hline
        Random & 
            {35.92} & 51.56 %
            \\
        RankGPT & 37.53 & 51.45 %
        \\
        RankGPT (reversed) & 32.98 & 44.67 %
        \\
        \rowcolor{gray!10} \ours & 
            {44.30} & 63.11 %
            \\
        \rowcolor{gray!10} \ours (reversed) & 26.26 & 39.81 %
        \\
        \thickhline
    \end{tabular}
    \caption{
    Not only the debiased ranking found by \ours leads to better performance with large margin, it exhibits higher polarity, incurring notable performance degradation when the ordering is reversed.
    }
    \label{tab_reverse}
\end{table}

\begin{table}[t]
    \centering
    \begin{tabular}{lrr}
        \thickhline
         & \multicolumn{2}{c}{\multirow{1.25}{*}{HotpotQA}}
         \\
        \cmidrule(r){2-3}
         \multirow{-1.25}{*}{Ranking} 
         & \multicolumn{1}{c}{\multirow{-1.25}{*}{EM}}
         & \multicolumn{1}{c}{\multirow{-1.25}{*}{GPT-4}}
           \\\hline
        Random & 53.05 & 78.89 %
        \\
        Self-consistency & 54.68 & 80.79 %
        \\
        \rowcolor{gray!10} \ours & 
            \textbf{56.65} & \textbf{82.27} %
            \\
        \thickhline
    \end{tabular}
    \caption{
    Results on HotpotQA with LLaMA-3 70B as the backbone LLM.
    }
    \label{tab:larger}
\end{table}

\begin{figure}[t]
  \centering
  \includegraphics[width=.9\linewidth]{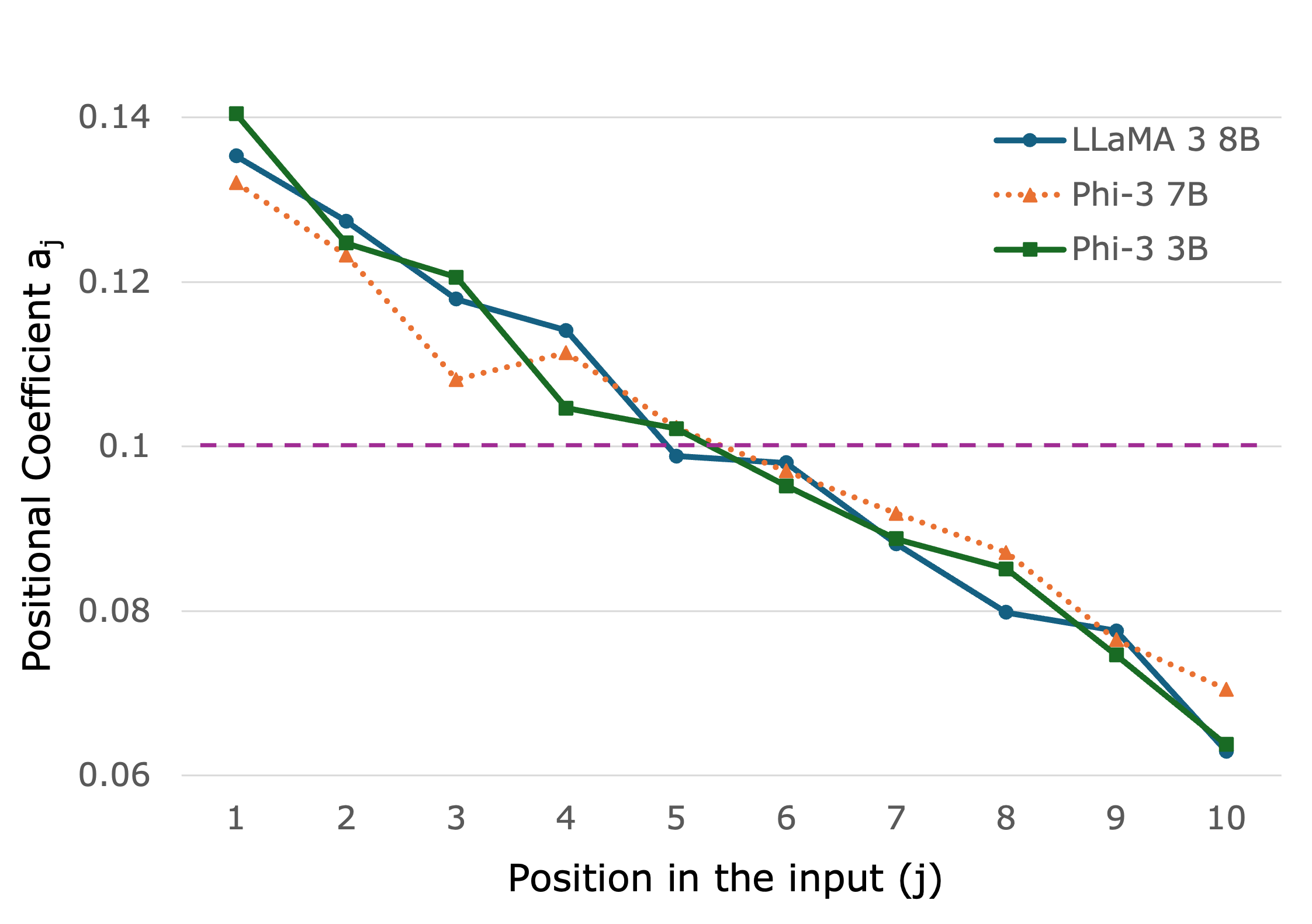}
  \caption{The values of computed positional coefficients $a_j$'s for each position $j$, averaged across datapoints for different models on HotpotQA. Dashed violet line represents the ideal case of zero position bias.}
  \label{fig:empirical_aj}
\end{figure}

\paragraph{Downranking gold if desirable} 
If there is no bias, ranking gold higher should optimize RAG output.
In contrast, if there is bias,
downranking a relevant passage, %
and an effective debiasing algorithm should identify downrankings that
may improve output accuracy.
Table~\ref{tab:goldrank} shows that, in the rank determined by \ours, the gold passage
does not necessarily surface higher, yet this still results in the generation of more accurate answers. The performance of \ours and baseline methods is also compared to scenarios where gold passages are consistently placed in certain positions; notably, \ours outperforms methods that achieve similar average gold passage rankings.

\paragraph{Quantified position bias} Figure~\ref{fig:empirical_aj} illustrates the average values of positional coefficients $a_j$ across different models, showing a monotonically decreasing trend as the passage's position moves from the beginning to the end of the prompt.
This quantifies a significant position bias, showing earlier passages contribute more to the final generated output.

\paragraph{Optimality of ranking}
In line with the spirit of studying the `reversal curse,'
suggesting 
LLM's ability to process reversed inputs would drop significantly when the original input order is desirable~\citep{berglund-etal-2024-reversal-iclr}, we explored reversing the ranking of contexts. Our findings reveal a significantly greater performance drop with our method compared to baseline approaches.
As shown in Table~\ref{tab_reverse}, reversing the sequence identified by \ours results in an 18-point drop in EM,  while for RankGPT, the decrease is less than 5 points.
This shows the ranking identified by \ours through intervention is ideal, such that
adversarially perturbing by reversing the rank
would 
harm the performance greatly.

\paragraph{Effect of model scale}
We provide evidence that larger models still suffer from position bias and can  benefit from applying \ours as well.
As shown in Table~\ref{tab:larger}, we observe consistent results with LLaMA-3 70B as the backbone LLM.

\section{Conclusion}

We proposed \ours, a novel inference-time scaling method for bridging the retriever and generator in RAG.
By modeling the position bias of LLMs from aggregated observations over multiple  interventions, \ours disentangles the impact of position from utility, enabling it to determine a debiased ranking of the contexts. We also demonstrated that leveraging the retriever's prior knowledge can reduce the search space of permutations, lowering both the number of LLM calls and the cost of each call.
Finally, we showcased the effectiveness of \ours across several benchmarks in question answering and other RAG tasks.

\section*{Limitations}
While we have presented results with LLaMA-3 70B as the generator in Section~\ref{sec:analysis},
experiments with more capable and sophisticated models are further needed to deepening our understanding of the sensitivity to input ordering of LLMs in RAG.

In addition, the proposed method increases inference compute usage as it invokes multiple forward passes for intervention.
However, there are many scenarios in which improving the performance is of more critical consideration than saving inference compute, e.g., healthcare.
We also discussed budget-constrained scenarios, for which we reduce both the number and latency of invocations in Section~\ref{sec:method_elaborate}.

Meanwhile, in extreme scenarios where only one invocation is allowed, intervention can be moved to training time, replacing inference cost with training compute for similar gains. This can be promising future directions and we leave it as next work.

\bibliography{anthology_part1,anthology_part2,custom}

\clearpage

\appendix

\begin{figure*}[t]
  \centering
  \includegraphics[width=.7\linewidth]{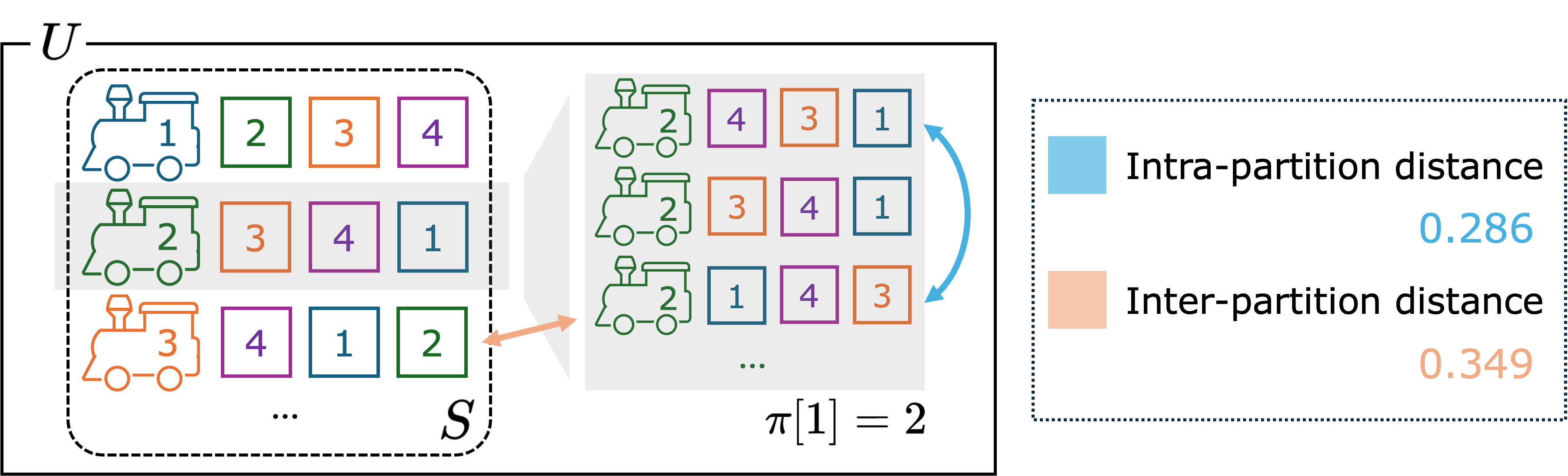}
  \caption{The set of all permutations $U$ can be partitioned into disjoint subsets based on the first item.
  Distance between two permutations can be measured by the L1 distance between the generator's predicted probability distribution on the first token of the response.
  Permutations from the same partition exhibit smaller distance in between in average, compared to permutations from different partitions.
  }
  \label{fig:l1_distance}
\end{figure*}

\section{Implementation Details}
\label{app:implementation}

For main experiments we have used LLaMA 3 8B Instruct, Phi-3 mini (3B) and small (7B) models available on huggingface as backbone models.
As mentioned in Section~\ref{subsec:exp_setting}, we have employed greedy decoding for generating the answer.

For preference distillation, we annotated about 20k examples in HotpotQA train set using the teacher model, scoring $K=30$ random permutations of the passages per query to build an offline preference dataset.
The student model, Phi-3 3B, was trained with LoRA at bf16 precision.
The relevant hyperparameter configuration was as follows:
for LoRA related settings, we used rank of $r=8$, $\alpha=32$, and dropout 0.1.
For general configuration, we used learning rate of 1e-4, effective batch size of 4; we trained the model for 5 epochs with weight decay of 0.01 applied.
We did not conduct hyperparameter search to determine these values, which leaves further rooms for improvement by performing one to find a better recipe.

Preference distillation did not introduce any degenerating behavior to the student model, such as a notable drop in QA performance.

\section{Comprehensive Sampling}
\label{app:comprehensive}
To argue comprehensiveness of $S$ more formally, we formalize cyclic permutations for $N$ passages to refer to the following set of permutations
\begin{equation}
S = \left\{ \phi^{(k)} \,\middle\vert\, 1\leq k \leq N \right\}
\end{equation}
where $\phi$ refers to some referential ordering 
\begin{equation}
\phi = \phi^{(1)} = [p_1, \cdots, p_{N} ]
\end{equation}
and 
$\phi^{(k)}$ denotes a permutation in which passages are shifted left by $k-1$ so that $p_k$ is placed at the beginning, that is,
\begin{equation}
\phi^{(k)} = [p_{k}, p_{k+1}, \cdots, p_N, p_1, \cdots, p_{k-1} ]
\end{equation}
for $k>1$. For example, a cyclic permutation by $2=3-1$ position to the left would give
\begin{equation}
\phi^{(3)} = [p_3, p_4, p_5, p_1, p_2]
\end{equation}
for $N=5$.

Here, the mapping $\mathcal{M}:U\rightarrow S$ from any permutation $\phi$ in $U$ to an element in $S$ is given as
all permutations starting with the same
passage.
This divides $U$ into $N$ non-empty and disjoint subsets, each of which maps
to $\phi^{(1)}, \ldots, \phi^{(N)}$, respectively.

Figure~\ref{fig:l1_distance} illustrates these concepts again as in Figure~\ref{fig:comprehensive_subset}, in which the permutations starting with passage 2 as the first item are all mapped to $\phi^{(2)}$ to form the subset $S$.
In order to confirm the common finding from previous literature that the first passage exerts the highest influence on generation, we show that the average distance between (two) permutations is closer in each partition, than between partitions.
The distance between two permutations $\pi_1$ and $\pi_2$ was measured by the L1 distance between two probability distributions, namely the generator's prediction on the first token of the response given the permutations:
\begin{equation}
d(\pi_1, \pi_2) = \sum_{y_1\in\mathcal{V}} \lvert P(y_1 \,|\, \pi_1) - P(y_1 \,|\, \pi_2). \rvert
\end{equation}
This distance captures how similar the model prediction would be given two different permutations of the same set of passages, suggesting that a permutation close to another can replace it without altering the generator's prediction greatly.

\section{Details of Baseline Methods}
\label{app:baselines}
We have considered the following baseline methods to assess the effectiveness of \ours as a bridge between the retriever and the generator, most of which aim to rerank the passages using pointwise or listwise signals from the generator.
Other than RankGPT, we have reimplemented each baseline's score computation, of which validity can be ensured from retrieval results such as in Table~\ref{tab:goldrank}.
Names are provided in accordance with those in Table~\ref{tab_main}.
\begin{itemize}
\item `Retriever' uses the initial ranking from the retriever. For some benchmarks such an ordering is unavailable.
\item RankGPT~\citep{sun-etal-2023-chatgpt} asks an LLM to sort the passages in descending order of relevance to the query. GPT-4 (\texttt{gpt-4o}) was used as the backbone for this ranking purpose. We directly used their code\footnote{\href{https://github.com/sunnweiwei/RankGPT}{\texttt{github.com/sunnweiwei/RankGPT}}} for running experiments.
\item Bayes saliency~\citep{merth2024superposition} uses the Bayes saliency score defined in Eq~\ref{eq:bayes_score} to rank the passages. Originally, the score was used to prune irrelevant contexts.
\item Bayes saliency +~\citep{merth2024superposition} uses the Bayes saliency score computed iteratively as in Eq~\ref{eq:bayes_score_iterative} to rank the passages.
\item QG~\citep{sachan-etal-2023-questions} uses the probability the model assigns to the query conditioned on each passage as the score, i.e., $u_p = P(q \,|\, p)$.
\item LongLLMLingua~\citep{jiang-2023-longllmlingua} defines an importance score per passage as the sum of the following token-level score over the tokens in the query condition:
\[ u_p = \sum_{l} P(q_l \,|\, p; q_{<l}) \log P(q_l \,|\, p; q_{<l}). \]
\item Self-consistency~\citep{selfconsistency} considers 30 random permutations of the retrieved passages to generate 30 answers, and chooses the answer most frequently appeared.
For comparison, we also reported the average score over those permutations, denoted as `Random.'
\end{itemize}

These baselines are indeed stronger as retrievers, as shown by their retrieval accuracy (MRR) previously presented in Table~\ref{tab:goldrank}.
It supports our key finding that stronger retrieval performance
does not necessarily lead to higher generation quality, when compared to the standard approach of using retriever-produced rankings on MS MARCO and CRAG.
This is consistent with the findings from \citet{cuconasu2024power} that adding noise to the retriever, which would make it `weaker' as a retriever, may lead to improvements in generation quality.
Our contribution is optimizing interventions towards
bridging retriever and generator.

\begin{table}[t]
    \centering
    \begin{tabular}{lr}
        \thickhline
         Metric
         & Score
           \\\hline
        Kohen's $\kappa$ & 
            {.874}
            \\
        Kendall's $\tau$ & 
            {.828}
            \\
        Fleiss' $\kappa$ & 
            {.694}
            \\
        \thickhline
    \end{tabular}
    \caption{
    Agreement between human annotators and human-LLM judgment.
    }
    \label{tab:humaneval}
\end{table}

\section{Soundness of GPT-4 Evaluation}
\label{app:humaneval}

We conducted a small-scale human study on the soundness of evaluation using GPT-4 and obtained results showing GPT-4’s evaluation is indeed highly correlated to human judgment as presented in Table~\ref{tab:humaneval}.
For obtaining Table~\ref{tab:humaneval}, 3 annotators were tasked with classifying 100 samples from MS MARCO as correct or incorrect, by comparing model generated responses against the ground truths.
We report the agreement between this human judgment and GPT-4's evaluation used in our paper, alongside the inter-annotator agreement, where strong agreement is indicated in all cases.
Human-GPT-4 agreement was measured by Cohen’s kappa after majority voting and Kendall’s tau correlation after soft label aggregation, while the inter-annotator agreement was measured by Fleiss’ kappa.

\begin{table}[t]
    \centering
    \begin{tabular}{lrr}
        \thickhline
         & \multicolumn{2}{c}{\multirow{1.25}{*}{HotpotQA-dep}}
         \\
        \cmidrule(r){2-3}
         \multirow{-1.25}{*}{Ranking} 
         & \multirow{-1.25}{*}{EM} & \multirow{-1.25}{*}{GPT-4}
           \\\hline
        Random & 
            {45.79} & 71.67 %
            \\
        \rowcolor{gray!10} \ours & 
            \textbf{51.92} & \textbf{77.57} %
            \\
        \thickhline
    \end{tabular}
    \caption{
    QA performance on HotpotQA subset of queries with inherent sequential dependency between the decomposed subquestions.
    }
    \label{tab:hotpot_dep}
\end{table}

\section{Tasks with Natural Inductive Bias}
For tasks involving reasoning chains, the order the evidence appears may play an important role in the generation quality~\citep{chen-etal-2024-premise-icml}.
While we have showed that \ours also work well for multi-hop reasoning scenarios on HotpotQA, here we provide more detailed discussion regarding the compatability of multi-hop reasoning and \ours.

To this end, we first identified cases with dependency, that is, those define a specific `natural order' of two subquestions, or corresponding gold passages in HotpotQA.
As a proxy for categorizing dependent subquestions, we prompted GPT-4 to decompose each query into two subquestions then categorize those with dependencies.
We found that approximately 1/4 of HotpotQA queries were non-dependent, meaning the two subquestions could be answered in any order.
On the remaining subset with dependencies, our method still demonstrated significant improvements over the random baseline, as shown in Table~\ref{tab:hotpot_dep}, which suggests a degree of robustness with respect to dependency.
However, further investigation is needed for datasets with stronger dependencies than those observed in HotpotQA, such as questioning temporal dependencies, as well as for more accurate categorization of such scenarios.

\section{Qualitative Analysis}
\label{app:qualitative}

Table~\ref{tab:examples} shows an example of a winning case for \ours compared to baseline methods, where it upranks the gold passage to produce the correct answer.
Among all the winning cases in pairwise comparison with Bayes saliency or Retriever baseline in terms of GPT-4 evaluation, about 79\% fall into this scenario, mainly accounting for the performance gain with \ours.

In this example, none of the passages directly mentioned the entity `graduate marketers' as it appears on the query, but the proposed method successfully resolved it as `MBA graduates' rather than `marketing managers' or similar ones, thanks to its approach of producing a passage score that is aware of the whole context by considering several permutations to mitigate the position bias.
In contrast, the pointwise baselines which predicts the relevance of each passage to the query separately fail to prevent passages about entities like `marketing managers' ranked higher, leading to undesirable answers.
Also, pointwise baselines are more prone to noises in presentation of passages such as repetition, as evidenced by $p_6$ in this example, which consists of a near duplicate of a three-sentence chunk is ranked the highest by the Bayes saliency method.
After removing the repetition, the rank determined by the Bayes saliency method drastically changes, while still leading to an incorrect answer.

On the other hand, Table~\ref{tab:examples_2} shows another example where the proposed method wins while not necessarily upranks the gold passage.
Specifically, the proposed method
again avoids highlighting \emph{distracting} passages, that would lead the generator to producing wrongful answers by effectively considering the whole passages.
Due to its high lexical similarity, the distracting passage ($p_{10}$) was paid undesirable attention by the pointwise baseline.

\begin{table*}[t]
    \centering
    \begin{tabular}{p{\linewidth}}
        \thickhline
         \emph{Question:} How much would \emph{graduate marketers} make?
        \\\hline
        \emph{Answer from Ours:}
        (Correct) According to the passages, the average starting salary for MBA graduates can be as high as \textbf{\$110,000 to \$130,000} per year … \\
        \emph{Top-3 Passages from Ours:}\\
        $p_3$: At present, the average starting salary for \textbf{MBA graduates} can be as high as \textbf{\$110,000 to \$130,000} … \\
        $p_7$: A general brand manager will make about \$5,000 less than a package design manager, … \\
        $p_5$: The BLS reports that the upper 10 percent of \textbf{marketing managers} earned more than \$187,200, as did managers … \\\hline
        
        \emph{Answer from Bayes saliency:}
        (Wrong) According to the passages, \textbf{graphic designers} in the top 10 percent made more than \textbf{\$77,370}, and \textbf{marketing managers} … \\
        \emph{Top-3 Passages from Bayes saliency:}\\
        $p_6$: … \textbf{Graphic designers} in the top 10 percent made more than \textbf{\$77,370}, … \\
        $p_2$: \textbf{Marketing managers} earned a median annual wage of \$123,220 in 2013, … \\
        $p_7$: A general brand manager will make about \$5,000 less than a package design manager, ... \\\hline
        
        \emph{Answer from R-then-G:}
        (Wrong) According to the passages, \textbf{marketing managers} earned a median annual wage of \textbf{\$123,220} in 2013, … \\
        \emph{Top-3 Passages from retriever:}\\
        $p_1$: \textbf{Marketing managers} made salaries ranging from about \textbf{\$58,000 to \$166,000} per year in 2010, … \\
        $p_2$: \textbf{Marketing managers} earned a median annual wage of \textbf{\$123,220} in 2013, … \\
        $p_3$: At present, the average starting salary for \textbf{MBA graduates} can be as high as \textbf{\$110,000 to \$130,000} … \\
        \thickhline
    \end{tabular}
    \caption{An example from MS MARCO development set where ours produces the correct answer as it upranks the gold passage $p_2$. Models were provided with all the 10 passages to generate the answer, while due to space limit the top-3 of them are presented here. The subscript identifying each passage is the rank of the passage determined by the retriever, Bing search engine in this case.}
    \label{tab:examples}
\end{table*}

\begin{table*}[t]
    \centering
    \begin{tabular}{p{\linewidth}}
        \thickhline
        \emph{Question:} How (many) ounces in (a) cup?
        \\\hline
        \emph{Answer from Ours:}
        (Correct) \textbf{8} fluid ounces to a cup. \\
        \emph{Top-3 Passages from Ours:}\\
        $p_5$: ... The mark at \textbf{8} fluid ounces indicates 1 cup. For \textbf{8} fluid ounces, use a measuring cup. … \\
        $p_6$: \textbf{8} fluid ounces to a cup. This is liquid measure. However the 16 fluid ounces that make the pint … \\
        $p_7$: In the US, 1 cup = \textbf{8} fluid ounces (*not identical to the avoirdupois ounce which is weight) … \\\hline
        \emph{Answer from Bayes saliency:}
        \textbf{(Distractor: highly similar lexically, but semantic outlier in the list)} There are \textbf{0.12500000001479} cup in a ounce. \\
        \emph{Top-3 Passages from Bayes saliency:}\\
        $p_9$: This is a very easy to use ounces to cup converter. First of all just type the ounces (fl oz) value in the text field … \\
        $p_3$: If you asked about the ounce that is rougly 28 grams in weight, then you should realize that … \\
        $p_{10}$: There are \textbf{0.12500000001479} cup in a ounce.  ... \\
        \thickhline
    \end{tabular}
    \caption{Another example from MS MARCO where ours produces the correct answer, while maintaining the rank of the gold passage which is not included in the top-3 passages.}
    \label{tab:examples_2}
\end{table*}

\FloatBarrier\clearpage

\end{document}